\documentclass{article}
\usepackage{nips14submit_e,times}
\usepackage{hyperref}
\usepackage{enumitem}
    \usepackage{epsfig}
    \usepackage{amsmath,amsthm}
    \usepackage{amstext}
    \usepackage{amsfonts}
    \usepackage{amssymb}
\usepackage{graphicx}
\usepackage{algpseudocode}
\usepackage{algorithm}
    \usepackage{eucal}
    \usepackage{graphicx}
    \usepackage{subfigure}

\nipsfinalcopy

\newtheorem{definition}{Definition}[section]
\newtheorem{lemma}{Lemma}[section]

\newtheorem{theorem}{Theorem}[section]

\begin{document}
%

\title{Sequential Relevance Maximization with Binary Feedback}
%
%
%
%
%

%
\author{
Vijay Kamble\\
       EECS, UC Berkeley\\
       vjk@eecs.berkeley.edu
       \And
Nadia Fawaz\\
       Technicolor, Los Altos\\
       nadia.fawaz@technicolor.com
       \And
Fernando Silveira\\
	Technicolor, Los Altos\\
fernando.silveira@technicolor.com
}
\maketitle

\begin{abstract}
Motivated by online settings where users can provide explicit feedback about the relevance of products that are sequentially presented to them, we look at the recommendation process as a problem of dynamically optimizing this relevance 
feedback. Such an algorithm optimizes the fine tradeoff between presenting the products that are most likely to be relevant, and learning the preferences of the user so that more relevant recommendations can be made in the future. 

We assume a standard predictive model inspired by collaborative filtering, in which a user is sampled from a distribution over a set of possible types. For every product category, each type has an associated relevance feedback that is assumed to be binary: the category is either relevant or irrelevant. Assuming that the user stays for each additional recommendation opportunity with probability $\beta$ independent of the past, the problem is to find a policy that maximizes the expected number of recommendations that are deemed relevant in a session.


We analyze this problem and prove key structural properties of the optimal policy. Based on these properties, we first present an algorithm that strikes a balance between recursion and dynamic programming to compute this policy. We further propose and analyze two heuristic policies: a `farsighted' greedy policy that attains at least $1-\beta$ factor of the optimal payoff, and a naive greedy policy that attains at least $\frac{1-\beta}{1+\beta}$ factor of the optimal payoff in the worst case. Extensive simulations show that these heuristics are very close to optimal in practice. 

\end{abstract}

\section{Introduction}
Predicting the preferences of users in order to present them with more relevant engagements is a fundamental component of any recommendation system \cite{schafer1999recommender,resnick1997recommender}. Over the years, a wide variety of approaches have been proposed for this problem (see \cite{adomavicius2005toward} for a survey). These include content based approaches that rely on generating user and item profiles based on available data \cite{pazzani2007content,lops2011content}, collaborative filtering approaches \cite{resnick1994grouplens,herlocker2004evaluating} that recommend items based on similarity measures between users and/or items, and a combination of both \cite{balabanovic1997fab,burke2002hybrid}. In this paper, motivated by several settings of interest in which explicit feedback about the relevance of the recommendations can be received from the user on small timescales, we pursue a less studied approach (see \cite{shani2002mdp}) of modeling the recommendation process as a sequential optimization problem. Below are a few examples of such settings.
\begin{itemize}
\item Online retail: A user enters an online shopping portal to purchase an accessory, e.g. a watch. She is sequentially presented with various design choices and based on her feedback to these designs, the system adaptively presents recommendations that are more likely to be liked by her.
\item Online media-on-demand services: A user using an online music-on-demand service would like to find a new genre of music to listen to. Short sound-clips are played for her sequentially, and based on the feedback that she provides for these clips, the recommendation system seeks to adaptively find genres that are better suited to her tastes.

\item Advertising in online video: As video ads are inherently more disruptive of a user's attention, and thus potentially more valuable than sponsored search ads, there is a strong motivation for designing ad 
allocation mechanisms that take into account the relevance of these ads to the users. Services like YouTube and Hulu collect explicit feedback about the relevance of an ad after it is shown, and this feedback can be used to adaptively learn the preferences of the users and show more relevant ads.
\end{itemize}

We consider a model that is derived from cluster models for collaborative filtering (see \cite{breese1998empirical}) in which the history of user behaviors is compressed into a predictive model, where users are classified into `types' that capture the preference profile of the user. A typical recommendation generation algorithm dynamically observes user behavior and uses maximum-likelihood estimates based on this predictive model to choose products that are more likely to be relevant. Our approach replaces this maximum-likelihood estimation with a sophisticated optimization problem, in which the two conflicting goals of presenting the most relevant products  based on current predictions of user preferences, and learning the underlying type of the user so that more relevant engagements can be shown later, are concurrently optimized in a precise and systematic way.

Our model assumes that a user that enters the system is sampled from a probability distribution over a set of types that is a priori known to the system designer. Each type is associated with a string of  `relevance' ratings for the different categories of products. We focus on the simplest case in which this relevance rating is binary, i.e. the user considers a category of products either relevant or irrelevant. We assume that the number of recommendation opportunities available in a session is random, modeled as a geometric random variable arising from the assumption that the user stays for each additional opportunity with a fixed probability $\beta$ independent of the past. Under this setting we focus on the problem of adaptively maximizing the expected cumulative relevance of recommendations presented to the user during the session.

Our main contribution in this paper is the analysis of this sequential relevance maximization problem. 
At first glance, one can see that the optimal policy can be determined using a naive recursive algorithm. But as is typical of such algorithms, it is highly inefficient due to repetition of redundant work. The standard tool to solve such problems is dynamic programming, which turns these inefficient recursive algorithms into efficient iterative solutions. But unfortunately in our case, the state space for this program grows exponentially in the number of types and categories. Further, an efficient enumeration of these states is difficult.

We first derive certain key properties of the structure of the optimal policy using probabilistic interchange arguments. Using these properties, we provide an algorithm that strikes a balance between recursion and dynamic programming to solve for the optimal policy. Unfortunately, this algorithm still remains computationally prohibitive. Motivated by our structural results, we then propose and analyze two heuristic policies: a `farsighted' greedy policy that is easier to compute and a naive greedy policy that is analogous to the maximum-likelihood prediction performed by typical recommendation systems. We then prove that these policies are approximately optimal, i.e. they achieve a constant factor of the optimal payoff. We finally perform extensive simulations on random problem instances and we observe that these heuristic policies typically perform much better than that predicted by our worst case bounds.

\subsection{Related work}

The idea of posing the recommendation process as an optimization problem is not new. To the best of our knowledge, its earliest appearance in literature can be traced back to $\cite{bohnenberger2001policies}$, which proposed a decision-theoretic modeling of the problem of generating recommendations (on a palm-top) for a user navigating through an airport.  \cite{shani2002mdp} proposed a framework for modeling the sequential optimization problem in online recommendation systems as a Markov Decision Process (MDP)  \cite{puterman2009markov}. Their underlying formulation is quite general
and their focus is on defining and establishing this paradigm. The model that we consider on the other hand is more structured and our focus is on the analysis of the resulting optimization problem.

The sequential relevance maximization problem is closely related to Bayesian multi-armed bandit problems. In a multi-armed bandit problem (MAB), first introduced by Thompson in \cite{thompson1933}, a decision-maker faces a set of arms whose reward characteristics are uncertain and seeks to optimize the sequence in which they are pulled so as to maximize some long-run reward. In such problems one faces the tradeoff between exploration, i.e. learning the reward characteristics of the arms, and exploitation, i.e.  accumulating rewards by choosing good arms based on current estimates. These problems have been commonly studied under two distinct settings: \emph{Bayesian} and \emph{stochastic}, with a different set of analytical approaches used in each. Our model falls in the Bayesian setting  \cite{gittins1989,whittle1980}, in which an initial prior distribution is assumed over the parameters of a probabilistic reward generating model for each arm, and one performs Bayesian updates of these estimates as rewards are observed. One then solves the well-defined problem of maximizing either the long-term average or discounted cost. The standard solution tool in this case is dynamic programming. The \emph{stochastic} setting \cite{lai1985, lai1987} (also see \cite{bubeck2012} for a recent survey) does not assume any prior distribution over the parameters and one instead tries to find policies that
minimize the worst case rate at which losses relative to the expected reward of best arm (called `regret') are accumulated \cite{auer2002finite,auer2003}.  The focus is on characterizing this optimal rate. 

Most of the literature in these settings has focused on the case where the rewards of different arms are statistically independent. In the Bayesian case, a seminal result by Gittins  \cite{gittins1989} shows that the optimal policy dynamically computes an index for each arm independently of all other arms, and picks the arm with the highest index at each step. 
But in our case the relevance of different products are correlated through the hidden user type and hence it is a type of a Bayesian MAB problem with \emph{correlated} or \emph{dependent} arms. It is well known that the decomposition result of Gittins does not hold for this case. Over the years there has been sporadic progress in tackling this problem, with most papers focusing on specific models. $\cite{feldman1962}$ and $\cite{keener1985}$ analyze two-armed bandit problems in which reward characteristics of two arms are known, but which arm corresponds to which reward distribution is not known, which leads to a natural dependence between the arms. In this case, the general case of more than two arms still remains open. \cite{pandey2007} studies another version of the problem in which the arms can be grouped into clusters of dependent arms, in which case the Gittins decomposition result can be partially extended. Recently, \cite{mersereau2009} considered a specific model of a MAB problem with dependent arms, where they analyzed the performance of a greedy policy and derived asymptotic optimality results. These type of problems have recently also gained attention in the stochastic setting  \cite{auer2003,danihayeskakade,tsitsiklispaat} (also see \cite{srikant2011} for the case of binary rewards), although the formulations and techniques in that setting are very different. The broad conclusion from this body of work is that the correlation between arms can be exploited to achieve better regret rates.
 
Another important difference between MAB problems and our problem is that, since any product can be presented only once and since there is a finite number of products in any category, there is a bound on the number of times each `arm' can be pulled. Thus one cannot `exploit' an arm forever and is forced to experiment intermittently. In the special case when each category has a single product, our problem is also related to the \emph{active sequential hypothesis testing} problem \cite{chernoff1959,naghshvar2013}. In this problem, one seeks to speedily learn a hidden random variable by adaptively choosing a sequence of correlates to observe, with a cost for each observation. This formulation would have been appropriate if our objective was to quickly learn the user type without any concern for the relevance feedback. But since our goal is to optimize the latter, a different approach is necessary.

\subsection{Structure of the paper}
The structure of the paper is as follows. In Section \ref{sec:model}, we introduce our model and define the relevance optimization problem. Section \ref{sec:opt} is devoted to the analysis of this problem, in which we derive key structural properties of the optimal policy and finally present an algorithm to compute it. In Section \ref{sec:approx} we propose two policies which are easier to compute and prove that they are approximately optimal. In Section \ref{sec:sim}, we extensively simulate our two approximately optimal policies on randomly generated problem instances and compare their performance to the optimal policy. Finally Section \ref{sec:conclusion} summarizes our work and discusses extensions to our model. The proofs of all our results can be found in the appendix.

\section{Model}\label{sec:model}
We consider the setting of a user who enters an online system and is sequentially presented with products from different categories, with the goal of maximizing the number of relevant products presented to him before he eventually leaves the system. Assume that there are $L$ total products. The products are divided into categories, with each category representing a set of similar products. Let these categories be labeled as $j\in \{1,\cdots,H\}=[H]$. Each category $j$ has $L_j$ products. A given user considers some set of categories to be relevant to him and this set is not known a priori. The system designer elicits explicit feedback about the relevance of a product after it is presented. This feedback is obtained as an answer to an explicit question, is assumed to be binary, and takes value 1 (resp. 0) when the product is relevant (resp. irrelevant). We assume that this feedback is accurately provided by the user. A product cannot be presented more than once to the same user during the session. Hence the maximum number of products that can be shown is restricted to $L$.

We capture the uncertainty in the preferences of the user by assuming that the user is one of $N$ possible types and the actual type of the user is a latent random variable that is not observed at the beginning of the session. Let $X\in[N]$ denote this random variable. Let $p_{X}$ be the corresponding probability distribution. We assume that the system designer only knows this distribution $p_{X}$. For each user type $i$ and for each product category $j$, let $q^i_j \in \{0,1\}$ denote the fixed binary relevance feedback of the user of type $i$ to that category. The type of the user is not known, and so for each category $j$, we introduce a random variable $Y_j\in\{0,1\}$ which represents the binary feedback of a user for any product in that category. $p_X$ induces a probability distribution on $Y_j$: 
$$P(Y_j=1)=\sum_{i=1}^{N}q^i_jp_X(i).$$

\begin{figure}
\begin{center}
\includegraphics[width=3in,angle=0]{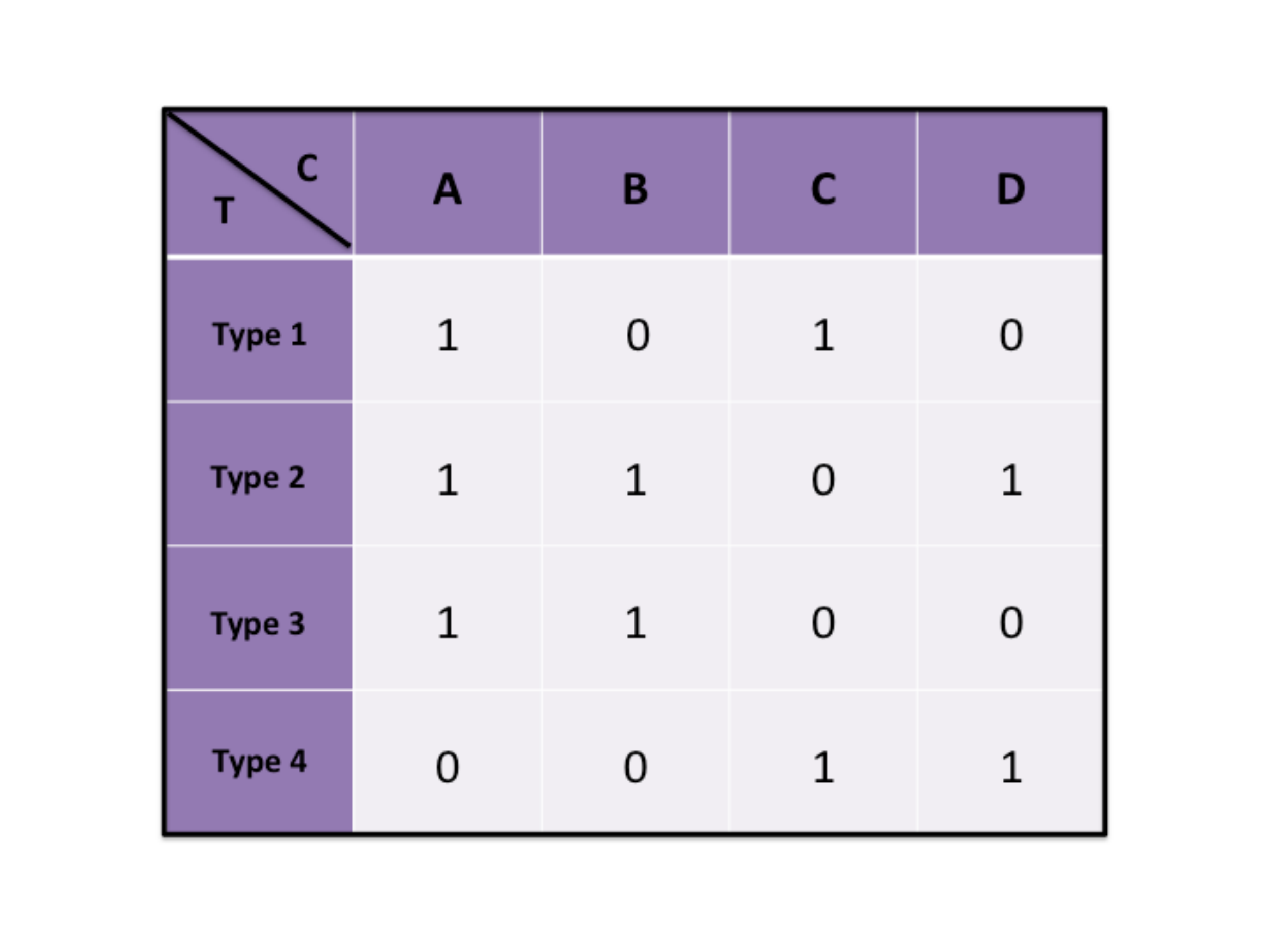}
\caption{A sample relevance matrix with 4 product categories (A,B,C,D) and 4 types (1,2,3,4) }
\label{fig:exam}
\end{center}
\end{figure}

It is convenient to associate each user type $i\in [N]$ with an $H$-length binary vector of the $\{q^i_j\}$, $j\in [H]$ values for different categories. Hence we can define a $N\times H$ \emph{relevance} matrix $Q=\{q^i_j\}$, whose rows represent user types, and columns represent product categories\footnote{Notice that type space is quite general. If for a user, there is a joint distribution over finding different categories relevant then we can think of a user as being a convex combination of the types corresponding to the realizations of binary relevance vectors with the associated probabilities. Also, if some variation is observed in the feedback received for products that belong to the same category, then each of the products can be declared as individual categories. Although this increases the problem size, our analysis remains applicable.}. Figure~\ref{fig:exam} is an example of a relevance matrix with four types of users labeled 1 to 4 and four product categories labeled A to D. Each category has some specified number of products. For instance, type 1 finds category A and C relevant and finds B and D irrelevant.

The number of display opportunities that are available before the user leaves the system is modeled as a random variable $C \in \{1,2,\cdots\}$ with a geometric probability distribution $p_C$ where $p_C(m)=\beta^{m-1}(1-\beta)$ for $m\geq 1$. In other words, the user dynamics in the system is modeled as a memoryless random process, in which a user stays for each additional opportunity with probability $\beta$ or exits with probability $1-\beta$, independently of the past. This assumes that at least one opportunity is always available. Finally, the random variable $C$ is independent of the user type $X$. 
The feedback for a product can be obtained after every display opportunity, but since the feedback for a product is the same for every other product in its category, one can assume that the feedback is requested and obtained only when the product presented belongs to a category that has not been shown before.

\subsection{Relevance maximization}

The primary objective of the system designer is to maximize the expected number of relevant products presented to a user in the session. Once a user enters the website, at each display opportunity, the system designer adaptively decides which product should be shown to the user, while taking all the user feedback obtained in the past into consideration. 
We define the objective formally.
A policy $\psi$ for the designer is the sequence of maps $\psi=\{\psi_1,\cdots, \psi_L\}$ where each map $\psi_t:H_t\rightarrow A_t$ is a mapping from the set of possible observations of user feedback until time $t$, denoted by $H_t$, to the set of possible actions $A_t$, which is the set of choices of products. Let $\Psi$ be the set of all feasible policies. The objective of the designer is to find a policy which maximizes the expected number of relevant ads shown in a session under the constraint that no product is shown more than once. Let $l_t$ denote the product chosen at time $t$. Once a policy $\psi \in \Psi$ is chosen, $l_t$ is a well defined random variable. With some abuse of notation, let $j(l_t)$ be its category. Then the objective of the publisher is the following.
$$\max_{\psi\in\Psi}E^{\psi}_{\{Y_{j(l_t)}\},C}[\sum_{t=1}^{C}Y_{j(l_t)}]$$
\begin{equation}\label{problem}
\textrm{ subject to }\sum_{t=1}^{C}l_t\mathbf{1}_{\{l_t=l\}}\leq 1 \textrm { for each product } l\in [L].
\end{equation}
Assuming memoryless user dynamics, the optimization problem~(\ref{problem}) takes the following form
\begin{equation}
\max_{\psi\in\Psi}\sum_{t=1}^{\infty}\beta^{t-1}E^{\psi}_{\{Y_{j(l_t)}\}}[Y_{j(l_t)}].
\end{equation}
As mentioned earlier, this problem is a type of a Bayesian multi-armed bandit problem with correlated rewards (see \cite{gittins1989,whittle1980}) with an additional constraint on the number of times each arm may be pulled. At a first look, one can solve this problem using the following recursive program in Algorithm 1. 
\begin{algorithm}
{\bf Algorithm 1 (Optimal)} Function $[\overline{V}(Q,p,\beta), A(Q,p,\beta)]$ where $Q$ is a relevance matrix and $p$ is a probability distribution over user types.
\begin{itemize}
\item If $Q$ is empty, return $\overline{V}(Q,p,\beta)=0$.
\item For a category $j$, let $M_j$ denote the set of user types which find $j$ relevant and let $P(M_j)=P(X\in M_j)$. Also let $Q^{j}$ be the matrix obtained after removing the column corresponding to category $j$ and the rows corresponding to all the user types in $M^c_j$ and let $Q_{res}^{j}$ be the matrix obtained after removing the column corresponding to category $j$ and the rows corresponding to all the user types in $M_j$. Finally, let $p^j$ denote the distribution on the user types conditional on the event $\{X\in M_j\}$ and $p^j_{res}$ be the distribution on the user types conditional on $\{X\in M^c_j\}$.
\item Then define 
 $$\overline{V}_j=P(M_j)\bigg(\frac{1-\beta^{L_j}}{1-\beta} + \beta^{L_j} \overline{V}(Q^j,p^j,\beta)\bigg) $$
 $$ +(1-P(M_j))\beta\overline{V}(Q_{res}^j,p_{res}^j,\beta)$$
\item Return 
$$\overline{V}(Q, p, \beta)=\max_j \overline{V}_j$$
$$A(Q, p, \beta)\in \arg\max_{j} \overline{V}_j$$
\end{itemize}
\end{algorithm}
But it is well known that such recursive algorithms can
be very inefficient. The usual problem is when recursion leads to repeating work.
This happens when you have overlapping subproblems, which is unfortunately
the case here. Turning these inefficient recursive
algorithms into efficient iterative algorithms is the role of dynamic programming.
This requires us to define a state space of possible `information states' for each opportunity $t$, which encapsulate all the information that has been gained till time $t$. In our case, the information state corresponds to a smaller relevance matrix obtained after computing the posterior distribution on the types, by eliminating all the rows corresponding to user types that have conditional probability 0 and all the columns corresponding to categories that have been exhausted. The state space thus grows prohibitively large with time and its enumeration is cumbersome. In the next section, we prove some structural properties of the optimal policy and based on these we provide an efficient algorithm that strikes a balance between recursion and iteration in order to compute this policy. These structural results are motivated by the following examples.

\begin{figure}
\begin{center}
\includegraphics[width=3in,angle=0]{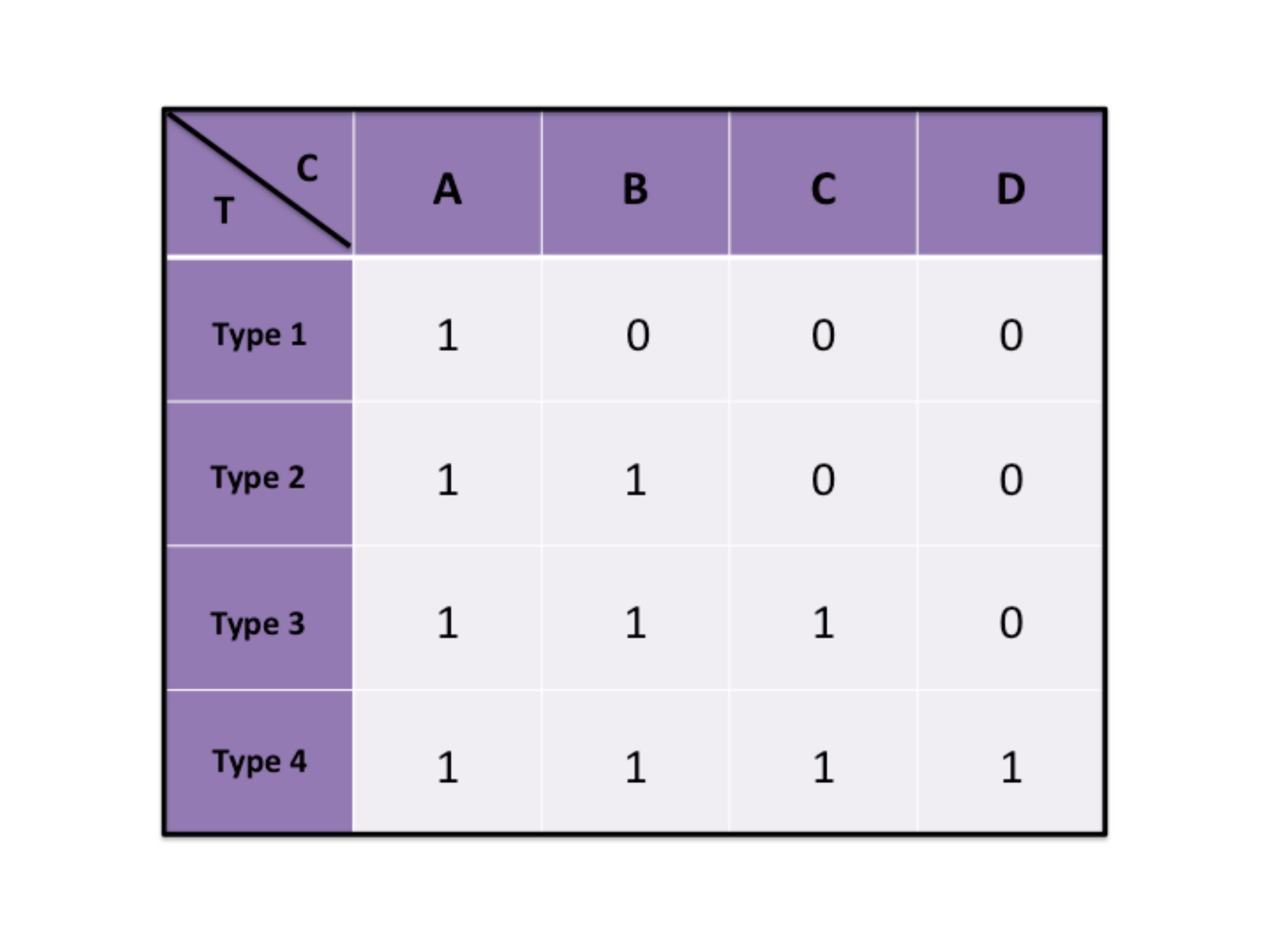}

\caption{A triangular relevance matrix. The optimal policy is to present categories in the order A, B, C and then D. }
\label{fig:tri}
\end{center}
\end{figure}

{\bf Example: A triangular relevance matrix :} Consider the relevance matrix shown in Figure~\ref{fig:tri}. A quick circumspection convinces us that the optimal policy is one which shows the categories in the order A, B, C and then D. If a positive feedback is obtained for a category then all the advertisers in that category are exhausted. To see this, observe that this policy attains the optimal payoff obtained in the case that the type of the user is known at arrival. 
Structurally, there is a partial order relation on the categories where one category `dominates' the other if the set of types which find it relevant is a strict subset of the set of types which finds the other relevant. This example shows that if this partial order relation leads to a complete ordering of the categories then the optimal policy simply presents the categories according to this order. But what if that is not the case? In lemma 2, we prove an appropriate generalization of this property for arbitrary relevance matrices using a probabilistic interchange argument. We show that if a category dominates some other then in the optimal policy it is presented before the other.

\begin{figure}
\begin{center}
\includegraphics[width=3in,angle=0]{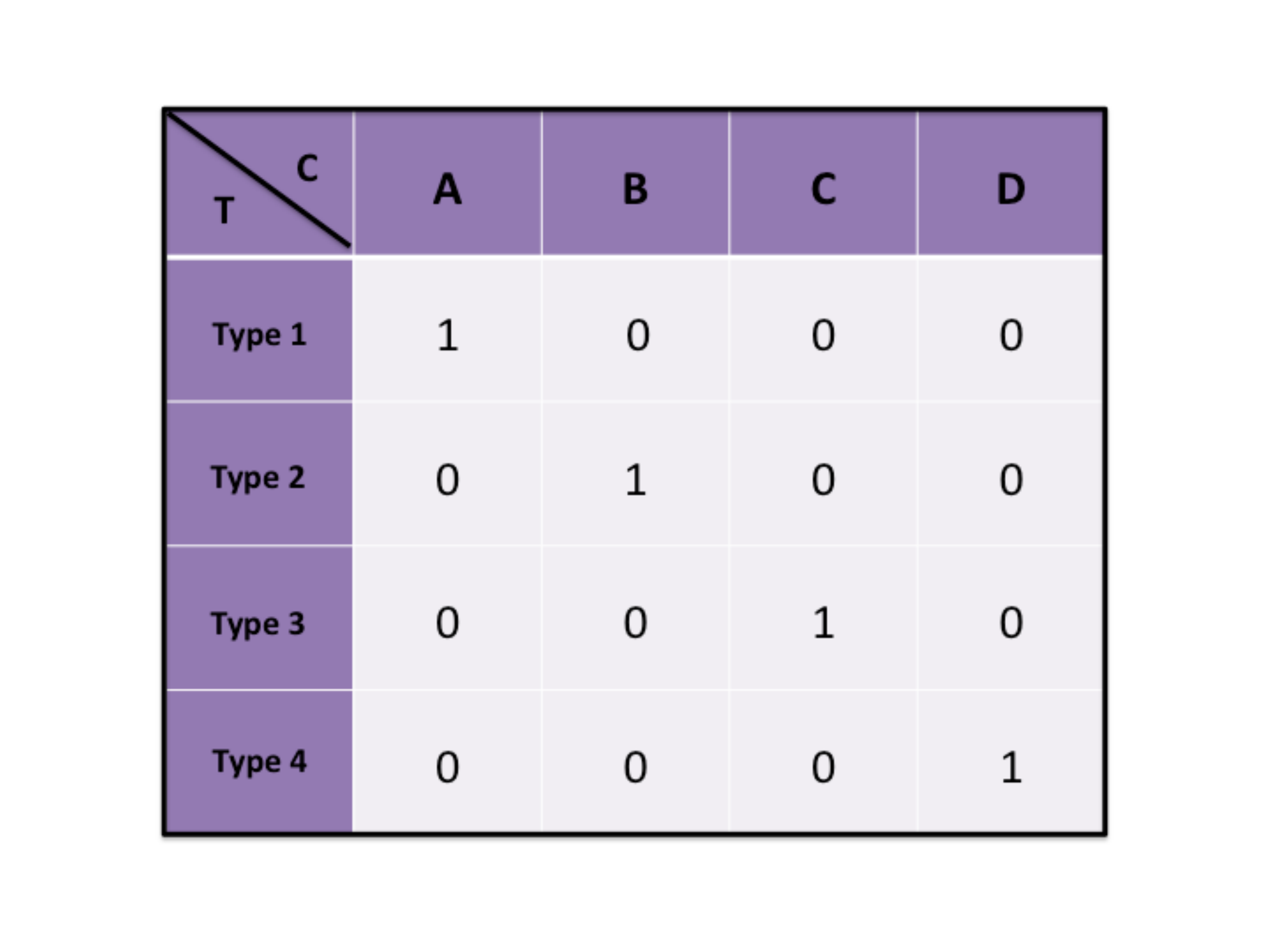}

\caption{A diagonal relevance matrix. The greedy policy is optimal. }
\label{fig:greedy}
\end{center}
\end{figure}

{\bf Example: A permutation relevance matrix :} Consider the relevance matrix shown in Figure~\ref{fig:greedy}. One can argue that in this case the optimal policy is greedy: choose the category with the maximum expected number of relevant ads. In fact, if the relevance matrix is a permutation of smaller block matrices, with multiple categories in each, we can consider the relevance optimization problem for each of the smaller blocks separately and greedily choose the order in which these blocks are chosen. 

\section{Characteristics of the optimal allocation policy}\label{sec:opt}

In this section we present some structural properties of the optimal allocation policy. 
\subsection{Property 1: If category A is relevant, show it}
We first present the following intuitive property.

\begin{lemma}\label{lem:positiveFBads}
In the optimal allocation policy, at any opportunity, conditional on the past observations, if there exists a product category $j$ that will generate a positive feedback with probability 1, i.e. $P(Y_j=1\mid H_t)=1$, then any product in $j$ that has not been shown is allotted immediately. If there are multiple such products then they can be allotted in any order.
\end{lemma}

This property implies that if a positive feedback is received for a product belonging to a particular category $j$, then all $L_j$ products of that category are scheduled to be presented in the immediately following opportunities\footnote{In order to not bore the user, we can introduce a bound on the number of products of the same category that can be successively shown to the user.}. The proof uses a simple probabilistic interchange argument.

\subsection{Property 2: If `likes A' implies `likes B', then show B before showing A}
To describe this next property, we first formally define a few ideas. In the dynamic allocation of products to the opportunities, we call an opportunity $t$ to be an \emph{experimentation} opportunity if conditional on information obtained until time $t-1$, there is not a single category $j$ such that $Y_j=1$ with probability 1. If there existed such a category, the previous lemma tells us to exhaust all the advertisers in that category. But since there is no such category, an experimentation opportunity brings to us the non-trivial problem of deciding which category to present to the user next. Thus all the non-trivial decisions in the optimal dynamic allocation policy are taken at the experimentation opportunities. Let $S(t)=\{i\in [N]: P(X=i\mid H_t)>0\}$
 be the set of user types that have a non-zero probability conditional on the history.
Then note that after observing the feedback from the allocation made at an experimentation opportunity $S(t-1)-S(t)\geq 1$. Let $E(t)$ be the set of categories available i.e. which have not been presented till opportunity $t$. Let $Q(t)$ be the relevance matrix with rows corresponding to the types in $S(t)$ and the columns corresponding to the categories in $E(t)$. Finally, for each category $j$ in $E(t)$, let $M_j(t)=\{i\in S(t): q^i_j=1\}$, which is the set of user types in $S(t)$ which find category $j$ relevant.  %
\begin{definition}\label{def:dominates}
We say that category $j$ \emph{dominates} category $j'$ at opportunity $t$ if $M_{j'}(t)\subset M_j(t)$. The categories that are not dominated by any other category are called non-dominated categories.
\end{definition}
For instance in Figure~\ref{fig:exam}, $A$, $C$ and $D$ are the only non-dominated categories since $A$ dominates $B$. Then we show the following.


%
\begin{lemma}\label{lem:undominated}
In the optimal allocation policy, at any experimentation opportunity, the product presented must be of a non-dominated category.
\end{lemma}
In other words, this lemma says that if the set of user types which find category A relevant is contained in the set of user types which find category B relevant, then in the optimal policy, category B is presented before category A. The proof of this lemma also uses a probabilistic interchange argument. Observe that the claim in the lemma is not an intuitively obvious fact. One may argue that in some cases, presenting a category that is dominated may help us learn the true user type faster and thus perform a better allocation in the future opportunities. Indeed if the goal is to minimize the expected number of opportunities taken to learn the user type exactly, then this property clearly does not hold (e.g. presenting a category that every user type finds relevant gives no information about the true type). 

Now let $U(t)$ be a generic class of non-dominated categories that satisfy the condition that $M_j(t)=M_{j'}(t)$ for all $j,\,j'\in U(t)$. This means that $U(t)$ is a class of categories found relevant by exactly same set of types. $U(t)$ will be called a \emph{non-dominated equivalence class} of categories and $M_{U(t)}$ denotes the set of types which find the class $U(t)$ relevant. We allow for a class to be singleton in the definition and so suppose there are $K(t)$ such non-dominated equivalence classes $\{U_1,\cdots,U_{K(t)}\}$ that partition the set of non-dominated categories in the relevance matrix. Let this set of non-dominated equivalence classes of categories be denoted by $\mathcal{U}(t)$. If furthermore the sets of types $\{M_{U_1},\cdots,M_{U_{K(t)}}\}$ are mutually disjoint, then we say that the set of non-dominated equivalence classes partition the type space. In this case, the relevance matrix can be represented as a block diagonal matrix composed of $K(t)$ smaller block matrices (up to permutation of the $K(t)$ blocks), with each block matrix corresponding to an equivalent non-dominated class. Such a small block is composed of columns of all 1s, one for each category in the class, and columns corresponding to the categories that the class dominates.

As products are presented and we recompute the relevance matrix after each feedback, we may lose non-dominated categories or new categories may become non-dominated. Thus the set of non-dominated equivalence classes will change. But in the case where new categories are added to a class of non-dominated categories, we want to be able to identify the new class with the old class. This can be done since the categories in an equivalence class in the relevance matrix at the first display opportunity will continue to remain in the same class as long as they are non-dominated and they have not been presented.  Thus a class $U$ in subsequent display opportunities is identified by equivalence to the set of categories in $U$ at the first display opportunity. For example, in the relevance matrix in Figure~\ref{fig:exam}, as mentioned before $U_1=\{A\}$, $U_2=\{C\}$ and $U_3=\{D\}$ are the non-dominated categories at the first opportunity. Suppose $C$ is presented and a negative feedback is received. Then in the new relevance matrix obtained after deleting rows corresponding to type $1$ and type $4$, and column corresponding to category $C$, the only remaining non-dominated class is $\{A,B\}$. In this case we identify $\{A,B\}$ with $U_1$, which was the class that contained $A$ in the first opportunity. Similarly if you present A initially and get a negative feedback, then \{C,D\} is left as the only non-dominated equivalence class, which is a result of merging classes $\{C\}$ and  $\{D\}$. In this case the new class is identified with any of the original classes $U_2$ or $U_3$. This brings us to the following property of any relevance matrix that can be easily verified.

\begin{lemma}
\label{lem:core}
Consider a relevance matrix with an initial set of non-dominated classes of categories $\mathcal{U}$. Suppose that a category from a class $U\in \mathcal{U}$ is presented. Suppose that a negative feedback is received for this category, and consider the new relevance matrix obtained after deleting the rows corresponding to user types that find the presented category not relevant and the column corresponding to the presented category. Then the new set of non-dominated equivalence classes of categories $\mathcal{U'}$  satisfies $\mathcal{U'}\subset \mathcal{U}$.
\end{lemma}
Intuitively this is because, when a negative feedback is obtained at some opportunity $t$, the rows corresponding to the user types that provide positive feedback to the shown category get deleted and thus it cannot happen that a category that was dominated at opportunity $t$  becomes non-dominated at $t+1$. 
On the other hand, after a positive feedback, completely new non-dominated equivalence classes can appear in the new relevance matrix computed after the posterior update. For example if A is presented and a positive feedback is received, the new relevance matrix has positive probability only on types 1, 2 and 3. In that case, D is dominated by B and hence (\{B\},\{C\}) is the new set of non-dominated equivalence classes (they are not equivalent), where notice that $\{B\}$ appears as a new (singleton) class.

\subsection{Structure of the optimal policy} The lemmas \ref{lem:positiveFBads}, \ref{lem:undominated} and \ref{lem:core} reveal the following structure of the optimal policy. Beginning from a set of non-dominated equivalence classes of categories, these classes are presented in a certain order as long as we keep getting a negative feedback. If any class obtains a positive feedback in the process, then we present all the products in that class,  `zoom in' to the next level (eliminating all the other types from the relevance matrix) and restart with a new set of non-dominated equivalence classes. Utilizing this structure, the following Algorithm 2 computes the optimal payoff.

\begin{algorithm}
{\bf Algorithm 2 (Optimal)} Function $\overline{V}(Q, p, \beta)$ where $Q$ is a relevance matrix and $p$ is a probability distribution over user types.
\begin{itemize}
\item If $Q$ is empty, return $\overline{V}(Q, p, \beta)=0$.
\item If $Q$ is non-empty, enumerate the non-dominated equivalence classes of $Q$. Let them be $(U_1,\cdots,U_K)$. Calculate the number of products in each class, denoted by $L^k=\sum_{j\in U_k}L_j$.  
\item 
For each $k=1,\cdots, K$, and each $\pi \subset \{1,\cdots,K\}$ such that $k\notin \pi$, let $\omega(\pi,k)$ be the event $\{X\in S(\pi,k)\}$ where 
$$S(\pi,k)$$
$$=\{i\in N: q^i_j=0\, \forall\, j\in U_s\textrm{, } s\in \pi \,\textrm{and} \,q^i_j=1\, \forall\, j\in U_k\}.$$ 
$S(\pi,k)$ is thus the set of user types that find all the classes with labels in $\pi$ irrelevant, but find the class $U_k$ relevant.
Let $Q^{\pi}_k$ be the relevance matrix obtained from deleting all the rows corresponding to user types in $S(\pi,k)^{c}$ and all columns corresponding to the categories in the classes in $\pi$ and the categories in $k$. Finally, let $p^{\pi}_k$ be the probability distribution on the user types conditional on the event $\omega(\pi,k)$. Then define 
 $$V^{\pi}_k=P(\omega(\pi,k))\bigg(\frac{1-\beta^{L^k}}{1-\beta} + \beta^{L^k} \overline{V}(Q_k^{\pi}, p^{\pi}_k,\beta)\bigg).$$
\item Return 
$$\overline{V}(Q,p,\beta)=OPT\triangleq$$
\begin{equation}\label{dyneqn}
\max_{{k_1,\cdots,k_{K}}\in \sigma(1,\cdots, K)}V_{k_1}+\beta V^{k_1}_{k_2}+\cdots + \beta^{K-1}V^{k_1,\cdots,k_{K-1}}_{k_{K}}.
\end{equation}
Here $\sigma(1,\cdots, K)$ is the set of permutations of the $K$ non-dominated equivalence classes.
\end{itemize}
\end{algorithm}
The optimization problem~$(\ref{dyneqn})$, is defined on the domain of all the possible orderings of the non-dominated equivalence classes of categories in $Q$\footnote{Note that if for some order, conditional on a sequence of classes getting negative feedback, if some class that is next in the order is dominated, then that order will simply not be chosen as the optimal order.}. This problem can be solved more efficiently using dynamic programming as opposed to comparing all the possible $K!$ orderings. One can define the state of the program at step $r$ as the set of classes $\{k_1,k_2,\cdots, k_r\}$ that have been presented till step $r$. A substantial reduction in the state space comes from the fact that $Q^{\pi_r}_k$ for any $k$ does not depend on the order in which the classes in $\pi_r$ were presented, and hence the state of the program at any step needs to only remember this set. Thus the size of the state space is ${K \choose 1}+{K \choose 2} +\cdots + {K\choose K}=2^K.$ 

At each state at step $r$, in the worst case, $K-r$ number of sub-programs need to be called to evaluate the set of payoff-to-go corresponding to the classes that have not been presented, i.e. $\{\overline{V}(Q_k^{\pi_r}):k\notin \pi_r \}$
Thus at each level in the recursive program, the number of sub-programs that are called is exponential in the number of non-dominated equivalence classes at that level in the worst case. Considering this, we turn to find good heuristic policies that are easier to compute.

\section{Approximately optimal policies}\label{sec:approx}


\subsection{Policy 1: Farsighted Greedy}
Consider the optimization problem (\ref{dyneqn}) assuming that we have been given $\overline{V}(Q^{\pi}_k)$ for each $k=1,\cdots, K$, and each $\pi \subset \{1,\cdots,K\}$ such that $k\notin \pi$. Now suppose that instead of optimally solving (\ref{dyneqn}), we adopt the following `greedy' policy. We iteratively define
$$k^*_s=\arg\max\{V^{k^*_1,\cdots,k^*_{s-1}}_i: i\in \{1,\cdots,K\}\setminus \{k^*_1,\cdots,k^*_{s-1}\}\}$$
 for $s=1,\cdots,K$. This policy assumes that the payoff-to-go from the `next level' onwards is given. But since it is not, we recursively compute an approximation to this payoff-to-go by assuming that we will follow the same greedy strategy in all the subsequent levels of the optimization problem. Algorithm 3 computes the proposed policy and its payoff.

\begin{algorithm}
{\bf Algorithm 3 (Farsighted greedy)}: Function $\overline{W}(Q, p, \beta)$ where $Q$ is a relevance matrix and $p$ is a probability distribution over user types.
\begin{itemize}
\item If $Q$ is empty, return $\overline{W}(Q,p,\beta)=0$.
\item If $Q$ is non-empty, enumerate the non-dominated equivalence classes of $Q$. Let them be $(U_1,\cdots,U_K)$. Calculate the number of products in each class, denoted by $L^k=\sum_{j\in U_k}L_j$.  
\item Let the event $\omega(\pi,k)$ be as defined in Algorithm 2. Similarly define $Q^{\pi}_k$ and $p^{\pi}_k$.
\item Iteratively compute
\begin{equation}\label{approx}
k^*_s=\arg\max\{W^{k^*_1,\cdots,k^*_{s-1}}_i: i\in \{1,\cdots,K\}\setminus \{k^*_1,\cdots,k^*_{s-1}\}\}
\end{equation}
where
 $$W^{\pi}_k=P(\omega(\pi,k))\bigg(\frac{1-\beta^{L^k}}{1-\beta} + \beta^{L^k} \overline{W}(Q_k^{\pi},p^{\pi}_k,\beta)\bigg).$$
\item Return $$\overline{W}(Q, p, \beta)=W_{k^*_1}+\beta W^{k^*_1}_{k^*_2}+\cdots + \beta^{K-1}W^{k^*_1,\cdots,k^*_{K-1}}_{k^*_{K}}.$$
\end{itemize}
\end{algorithm}
Note the computational savings as compared to the algorithm for computing the optimal policy. The comparison in equation~(\ref{approx}) when $s$ classes have been presented already is over $K-s$ possibilities in the worst case. Thus the number of times a sub-program is called is $K+(K-1)+(K-2)+\cdots+1=\frac{K(K+1)}{2}.$
Thus at each level in this recursive program,  the number of sub-programs that are called is quadratic in the number of equivalence classes at that level. We can then prove the following performance guarantee for this policy.
\begin{theorem}\label{algo1}
Let $L_{min}=\min_{j\in 1,\cdots,H} L_j$ be the minimum number of products in any category and let $H$ be the total number of categories. The farsighted greedy algorithm achieves $\frac{1-\beta^{L_{min}}}{1+\beta-\beta^H-\beta^{L_{min}}}$ factor of the optimal payoff. \end{theorem}

Note that the worst case is when $H$ is large and $L_{min}=1$, in which case the adaptive greedy policy achieves a $1-\beta$ factor of the optimal payoff. The key idea of the proof is as follows. The departure from optimality at any level has two sources: the fact that the payoff-to-go from the next level onwards is an approximation to the optimal payoff-to-go, and the order in which the non-dominated classes are presented in the current level is chosen greedily. If one assumes that the ratio of the approximation to the optimal payoff-to-go and the optimal payoff-to-go at the next level is some $\gamma$, and if one can quantify the departure from optimality of the greedy policy at the current level, one can compute a bound for the worst case ratio of the current payoff-to-go and the optimal current payoff-to-go as some $\gamma'=f(\gamma)$. One can show that this operator is a contraction. Thus one can recursively find a sequence of lower bounds that are uniformly bounded below by the fixed point of this sequence, which is the quantity in the theorem.

Note that the description of Algorithm 3 can be simplified by fully exploiting its recursive structure; we presented it in the current form to show the correspondence to Algorithm 2 and also to facilitate the argument in the proof of Theorem \ref{algo1}. The equivalent implementation can be found in the appendix.

\subsection{Policy 2: Naive Greedy}
Another simple heuristic that we can use is the following greedy policy. 
\begin{algorithm}
{\bf Policy (Naive Greedy)}
Let the set of non-dominated equivalence classes at an experimentation opportunity  $t$ be $(U_1(t),\cdots, U_K(t))$, and $(L^1(t),\cdots, L^K(t))$ be the number of products in each of these classes. Then choose a product from a class $k^*$ where 
$$k^*\in \arg\max_{k\in \{1,\cdots, K\}} \frac{1-\beta^{L^k(t)}}{1-\beta}P(X\in M_{U_k(t)}\mid H_t).$$
\end{algorithm}

We then have the following theorem.
\begin{theorem}\label{algo2}
The Naive Greedy policy achieves $\frac{1-\beta^{L^{min}}}{1+\beta-\beta^H}$ factor of the optimal payoff.
\end{theorem}
Note that in the worst case, when $L^{min}=1$ and $H$ is large, the greedy algorithm achieves at least $\frac{1-\beta}{1+\beta}$ factor of the optimal payoff. The proof of this theorem is similar to that of Theorem \ref{algo1}.
\subsection{A lower bound for $\beta$ close to 1}
We can also obtain a lower bound on the ratio of payoffs under either of the heuristic algorithms and the optimal algorithm for values of $\beta$ close to $1$. Intuitively, this follows from the observation that if user stays for long enough so that the number of ad opportunities available is greater than  $L$, then any policy obtains all the positive feedback that one can possibly obtain. 

\begin{theorem}\label{thm:highbeta}
Any feasible policy attains $\beta^{L-1}$ factor of the optimal payoff.
\end{theorem}

\section{Simulations}\label{sec:sim}
\begin{figure*}
 \centering
 \subfigure[$5\times 5$ (average)]{
  \includegraphics[width=1.6in]{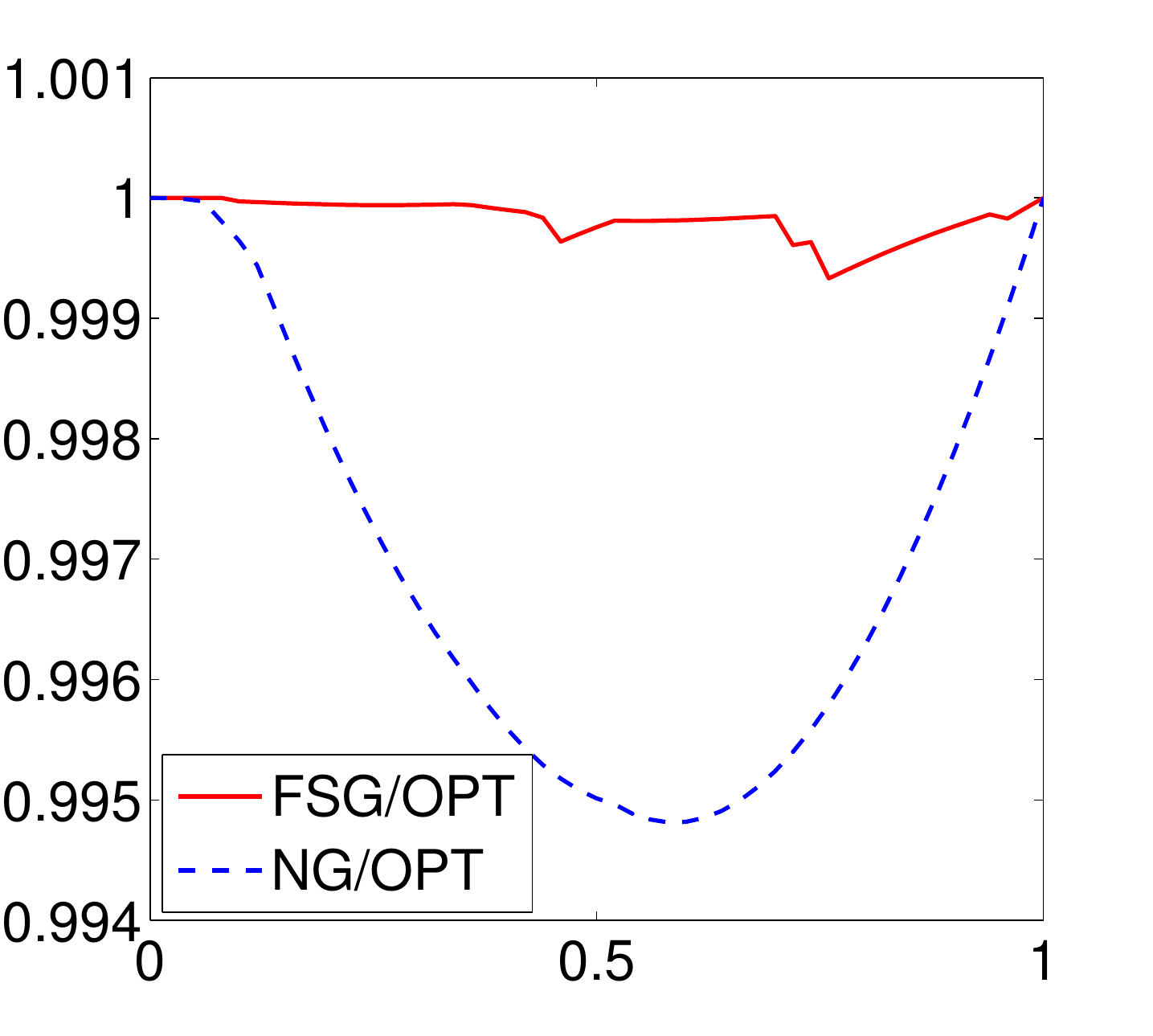}
   \label{fig:7avg}
   }
 \subfigure[$5\times 5$ (minimum)]{
  \includegraphics[width=1.6in]{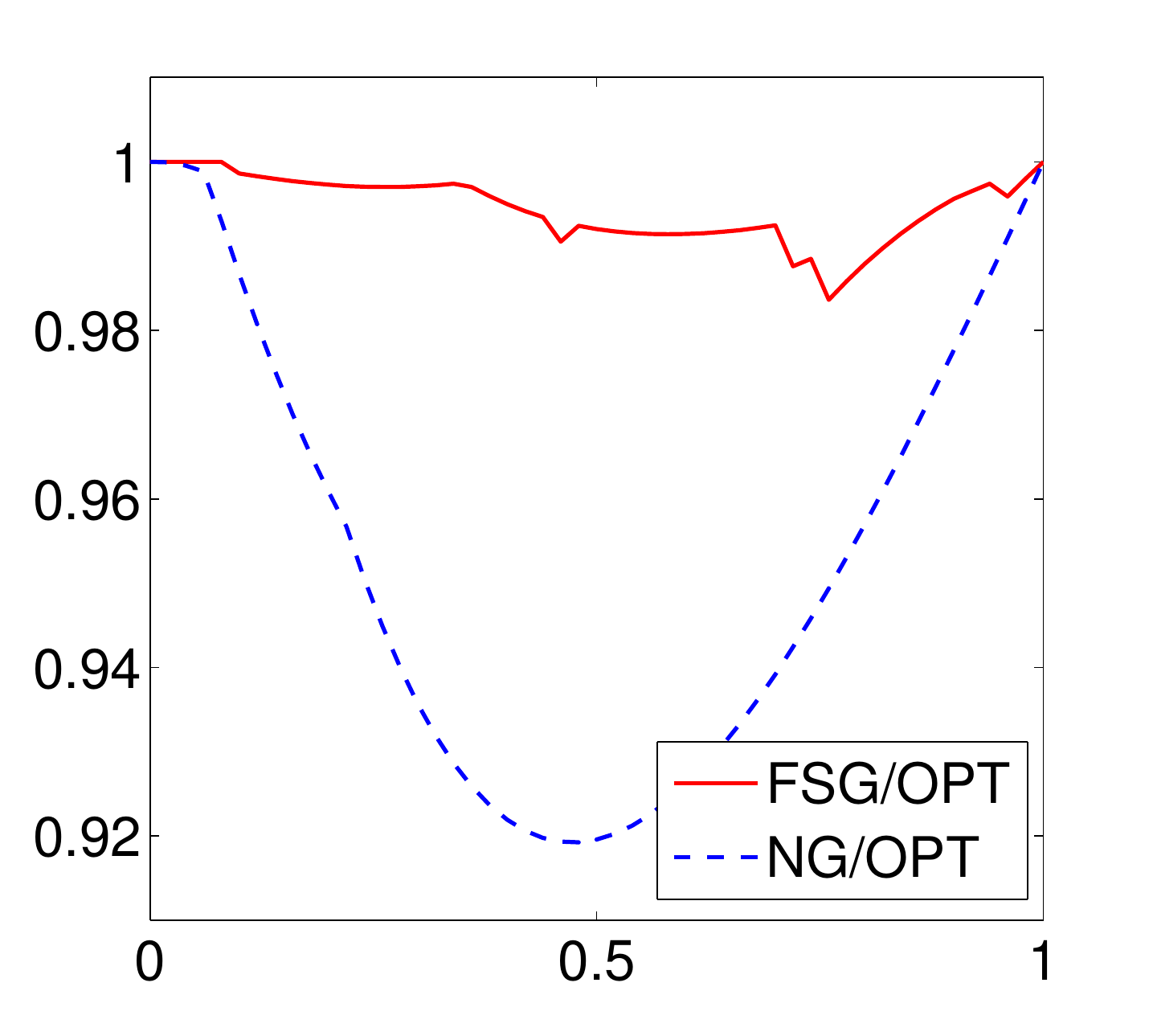}
   \label{fig:7min}
   }    
 \subfigure[$7\times 7$ (average)]{
  \includegraphics[width=1.6in]{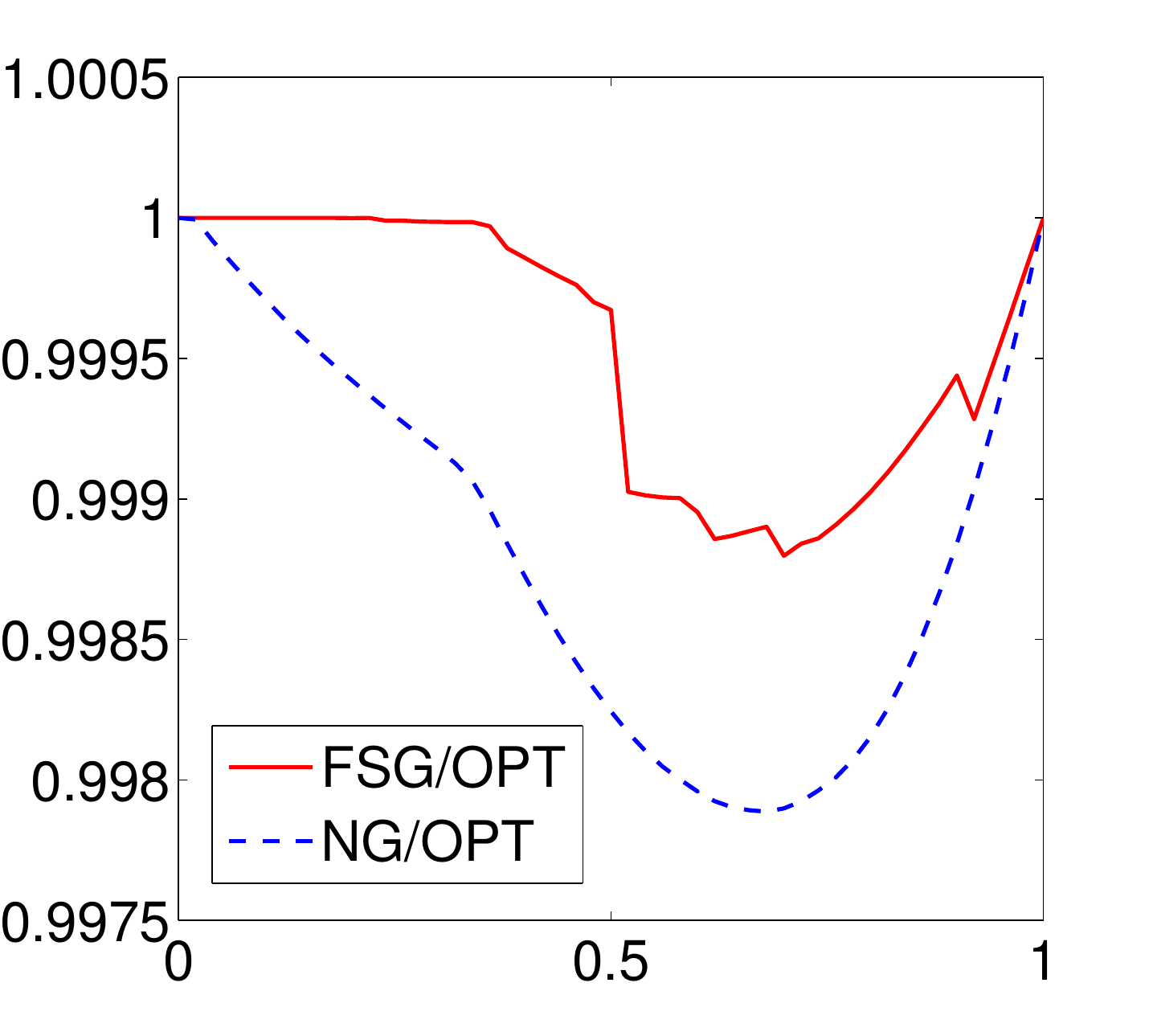}
   \label{fig:5avg}
   }
 \subfigure[$7\times 7$ (minimum)]{
  \includegraphics[width=1.6in]{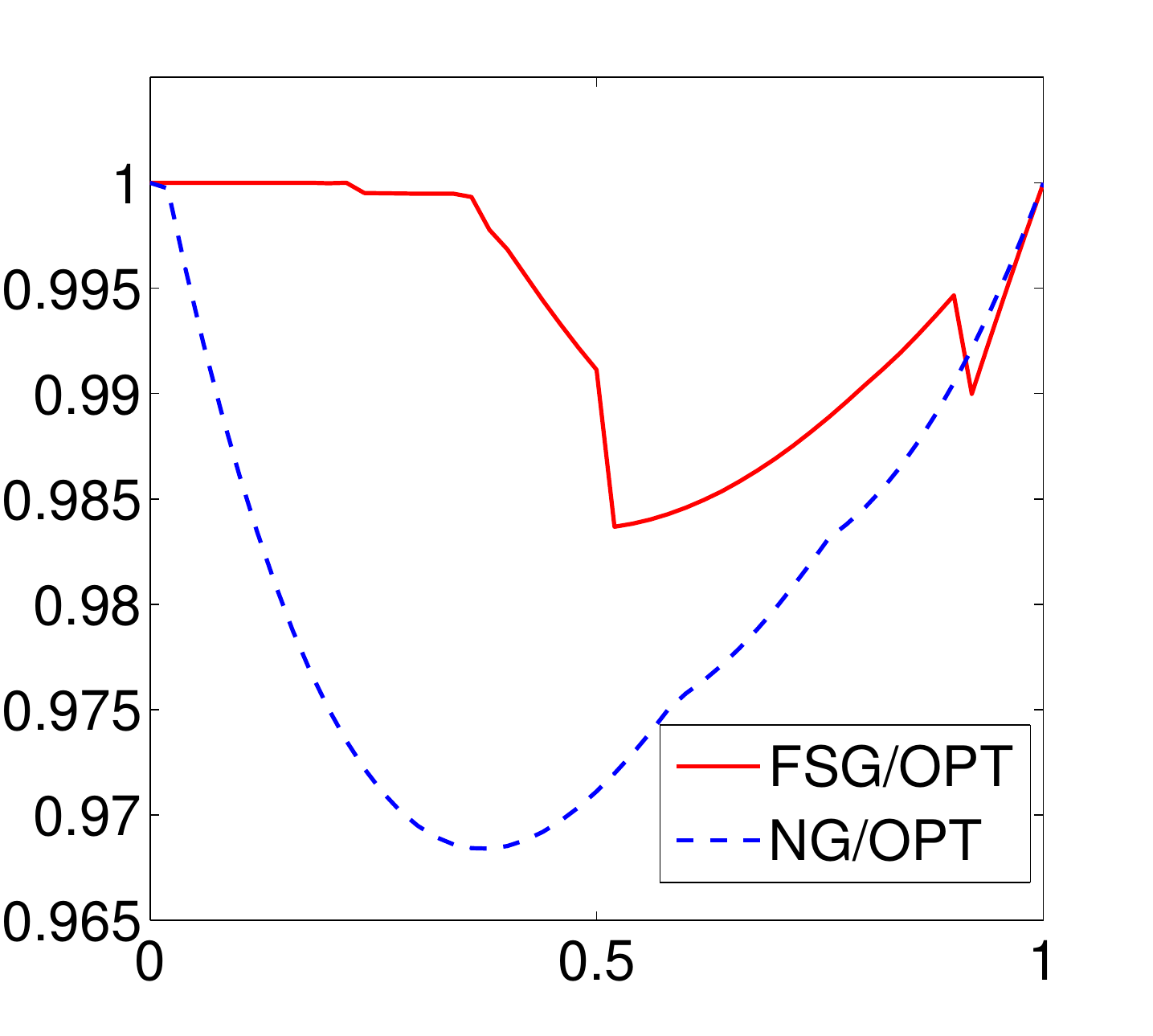}
   \label{fig:5min}
   }
    \caption{Average and worst case ratios of payoffs under either of the heuristic policies and the optimal policy across 50 samples.} 
   \label{fig:sim}
\end{figure*}

In this section we compare the performance of the the greedy with foresight policy and the naive greedy policy with the optimal policy. We generate 50 random samples each of $5\times 5$  and $7\times 7$ relevance matrices with associated randomly chosen priors. We compute the payoff under all the three policies, for $\beta$ ranging from $0$ to $1$. For each $\beta$, we then plot the average and the minimum  across the 50 samples of the ratio of the payoff under a non-optimal policy and the optimal policy. Our results are shown in Figure~\ref{fig:sim}.

Note that both the policies perform very close to optimal even in the worst case across the samples. Also, observe that for $\beta$ close to $0$ and for $\beta$ close to $1$, the payoff under both the policies approach the optimal payoff, which corroborates our bounds in theorems~\ref{algo1}, \ref{algo2} and~\ref{thm:highbeta}. The curve corresponding to the naive greedy policy is smooth because the policy does not depend on $\beta$ and hence the resulting payoff is continuous in $\beta$ (and so is the optimal payoff).

\section{Discussion and Conclusions}\label{sec:conclusion}


Our main contribution in this paper is the introduction and analysis of the sequential relevance maximization problem with binary feedback.  This problem naturally arises in several settings where a designer needs to adaptively make a sequence of suggestions to a user while learning his preferences from his feedback. This basic framework is amenable to extensions that adapt our approach to a more practical setting where some of our assumptions may not hold. For example, we assume that the number of display opportunities in a session is independent of both the type of the user and the relevance feedback, which may not hold in practice. For example, a user may be more likely to leave sooner if he is consecutively shown irrelevant products. 
Also, one of our central assumptions is that the user feedback is binary, but in practice one may benefit from a more fine-grained feedback from the user. For example, the user may convey a rating for the product which may be a number from 0 to 5. In this case, one would want to maximize the sum of ratings obtained for the products shown in a session. Another interesting extension is to incorporate the values of the products  so that one maximizes the total value of relevant products shown to a user in a session. We leave these extensions for future work. \\
{\bf User type and personalization:} In the current era of personalization of web services, it is important that a recommendation system be sensitive to the transience in the preferences of the users. For example, a user's preference for music can change every day, depending on her mood, company etc. A sequential optimization approach to generating recommendations can proactively learn these changes in user preferences by freshly eliciting relevance feedback for carefully chosen products, each time the user enters the system. 

In the model that we have considered in this paper, the type of a user captures her preferences for the session under consideration and the prior distribution over these types is assumed to be known to the system designer. One interpretation of this distribution is that it captures the preferences of a `typical' user in the population, and it is estimated from the observed behavior of all the past users. In another interpretation aligned with the notion of personalization, one can think of this distribution as capturing the variation in the preferences of the same user over multiple sessions. For example, in a naive interpretation, one can imagine that the type captures the `mood' of a person, which is sampled independently everyday, and her preferences for music on a particular day depends on her mood on that day. Even more generally, there could be cross-temporal dependencies in these types.  If one desires to optimize the performance of the recommendation process over multiple sessions, one needs to also estimate this type evolution process. We leave these considerations for future work.

\bibliography{references}
\bibliographystyle{plain}
\section{Appendix}

\subsection{Proof of lemma~\ref{lem:positiveFBads}}\label{ap:proofLemma1}
\begin{proof}
The result follows from a simple interchange argument. Suppose at time $t$, the posterior distribution over the set of possible types is $\{P(X=i\mid H_t) \}=(p^1_t,\cdots,p^N_t)$ and the set of remaining products is $A(t)$. Consider the event $W$ with a fixed realization of the user type $X=i$ and a fixed realization of the random variable $C=c$, which is the time at which the user leaves. Thus on this event, the string of binary feedback for the different categories is $\{q^i_j: j\in [H]\}$. Then for any fixed policy $\psi'$, the sequence of allocations of the products from time $t$ onwards till time $c$ is dictated by the policy and is determinate. Let this sequence of allocations be $\{l_t,l_{t+1},\cdots,l_c\}$ and the corresponding sequence of feedback be $\{y_{j(l_t)},y_{j(l_{t+1})},\cdots,y_{j(l_c)}\}$. Suppose there exists an advertiser $l^*\in A(t)$ such that, conditional on observations till time $t-1$, $Y_{j(l^*)}=1$ w.p. 1.
We now consider 2 cases:\\
{\bf Case 1 :} assume that on event $W$, for policy $\psi'$, $l^*\in \{l_t,l_{t+1},\cdots,l_c\}$. Say $l_{t'}=l^*$ for $t' \in \{t,\cdots,c\}$. Then if $t'\neq t$, we will construct a policy which generates the sequence of allocations $\{l^*,l_{t},\cdots,l_{t'-1},l_{t'+1},\cdots,l_c\}$ and thus give the same payoff on event $W$. This policy $\psi$ is the following:
\begin{enumerate}
\item Allot advertiser $l^*$ at time $t$.
\item from time $t+1$ onwards follow policy $\psi'$ assuming $H_{t+1}=H_{t}$
\item When $\psi'$ prescribes allotting $l^*$ at time $t'+1$, use the information $y_{l^*}=1$ to update the history to $H_{t'+1}$ and remove $l^*$ from the set of available advertisers. Follow the prescription of $\psi'$ for allotting advertisers from $t'+1$ and onwards.
\end{enumerate}
Clearly, this policy gives the same payoff on event $W$ since only the time at which $l^*$ is allotted has been interchanged. \\
{\bf Case 2 :} assume that on event $W$, for policy $\psi'$, $l^*\notin \{l_t,l_{t+1},\cdots,l_c\}$. Then observe that the policy $\psi$ described above generates sequence of allocations $\{l^*,l_t,l_{t+1},\cdots,l_{c-1}\}$. Thus the difference in payoff is given by
$$Y_{j(l^*)}-Y_{j(l_c)}=1-Y_{j(l_c)}\geq 0$$
Thus the number of relevant ads shown under policy $\psi$ is at least as high as that under policy $\psi$ for every such event $W$ from a set of disjoint events whose union is the entire probability space. Thus the expected number of relevant ads shown is also at least as high.
\end{proof}

\subsection{Proof of lemma ~\ref{lem:undominated}}\label{ap:prooflemma2}

\begin{proof}
The proof uses a probabilistic interchange argument. Suppose that at opportunity $t$, the optimal policy $\psi'$ allots a product $l'$ of category $j'$ while there exists a category $j$ such that $M_{j'}(t)\subset M_{j}(t)$. Let $l$ be any generic product of category $j$. Now consider a policy $\psi$ which allots category $j$ before category $j'$ by allotting product $l$ at opportunity $t$. To further describe this policy, we consider two cases:\\
\begin{enumerate}[label=(\Alph*)]
\item If the user finds $j$ relevant, it exhausts all the products in that category and then moves on to allotting $j'$. After it allots $j'$ it behaves as if $j$ was never allotted until $\psi'$ prescribes allotting $j$, upon which the designer updates the information that $j$ is relevant and moves on to allot the next category prescribed by $\psi'$ and so on.
\item If the user finds $j$ not relevant, then from time $t+1$ onwards it acts as if it allotted $j'$ at time $t$ and found that $j'$ is not relevant. Then when $\psi'$ prescribes allotting $j$, the designer updates the information that $j$ is not relevant and moves on to allot the next category prescribed by $\psi'$ and so on.
\end{enumerate}
We will show that on every disjoint event of the underlying probability space, the system designer shows at least as many relevant products by following policy $\psi$ instead of $\psi'$.
Consider an event $W$ on which the user leaves after opportunity $c\geq t$ and on which the realization of the user type is $X=i$. Then for the policy $\psi'$, the sequence of allocations of the products from time $t$ onwards until time $c$ is dictated by the policy and is determinate. Let this sequence of allocations be $\{l_t,l_{t+1},\cdots,l_c\}$ and the corresponding sequence of feedback be $\{y_{j(l_t)},y_{j(l_{t+1})},\cdots$ $,y_{j(l_c)}\}$. These allocations and feedback depend on the type $i$ that was realized on $W$. For this, we consider 3 mutually exclusive and exhaustive cases:\\\\
 {\bf Case 1:} We first consider the case where on the event $W$, $X=i\in M_{j'}(t)$. Then observe that $y_{j(l_t)}=y_{j(l')}=1$ and immediately the designer deduces that $Y_j=Y_{j'}=1$. Thus since $\psi$ is optimal, by the previous lemma w.l.o.g. the first $L_{j'}+ L_j$ allocations in the sequence $\{l_t,l_{t+1},\cdots,l_c\}$ can be assumed to be all the advertisers belonging to categories $j'$ and $j$ and thus the feedback is a sequence of $L_{j'}+L_j$ $1$s. Note that policy $\psi$ will be operating under case (A) and thus it will also generate a sequence of allocations in which the first $L_{j'}+L_j$ allocations in the sequence $\{l_t,l_{t+1},\cdots,l_c\}$ will be all the products belonging to categories $j'$ and $j$ (in a different order) and after that the rest of the sequence of allocations is identical to that under $\psi'$. Thus on such an event $W$ both the policies $\psi'$ and $\psi$ generate the same sequence of relevance feedback.\\\\
 {\bf Case 2:} Let us now consider the case where on the event $W$, $X=i\in M_{j}(t)-M_{j'}(t)$. In this case $y_{j(l_t)}=y_{j(l')}=0$ but $Y_j=1$. In this case the policy $\psi'$ generates the sequence of allocations $\{l',l_{t+1},\cdots,l_c\}$ and gets the feedback $\{0,y_{j(l_{t+1})},\cdots,y_{j(l_c)}\}$. Where as observe that the policy $\psi$ operates under case (1) and the designer discovers that $Y_j=1$ by allotting $l$ and then continues to exhaust all the products in $j$ before switching to the prescriptions of $\psi'$. Thus $\psi$ generates a sequence of allocations in which all the products in $j$ are allotted first and then the prescription of $\psi'$ is followed as described in case (A). In the case where $\psi'$ prescribed allotting $j$ at some opportunity and was able to allot all the products in $j$ until the final opportunity $c$, this leads to a sequence of allocations which is just a different ordering of the elements of the sequence $\{l',l_{t+1},\cdots,l_c\}$ and thus generates the same number of relevant products shown until time $c$. In the case where $\psi'$ allotted $0\leq r <L_j$ products of category $j$ up until the final opportunity $c$, then under policy $\psi$, the last $L_j-r$ products in the sequence $\{0,y_{j(l_{t+1})},\cdots,y_{j(l_c)}\}$ are dropped out in lieu of the same number of products in category $j$ in the beginning. But since all products in $j$ are relevant, this number of relevant ads under policy $\psi$ is still at least as high as that under $\psi'$.\\\\
{\bf Case 3:} Now consider the case where on the event $W$, $X=i\in S(t)-M_{j}(t)$. In this case $y_{j(l_t)}=y_{j(l')}=0$ and also $Y_j=0$. Thus the policy $\psi'$ generates the sequence of allocations $\{l',l_{t+1},\cdots,l_c\}$ and gets the feedback $\{0,y_{j(l_{t+1})},\cdots,y_{j(l_c)}\}$. In this case $\psi$ operates under case $(B)$. Now in the case where $l\not\in \{l',l_{t+1},\cdots,l_c\}$ for any product $l$ in category $j$, $\psi$ generates the same sequence of feedback $\{0,y_{j(l_{t+1})},\cdots,y_{j(l_c)}\}$. In the case where $l\in \{l',l_{t+1},\cdots,l_c\}$ for some product $l$ in category $j$, then under $\psi'$, since $j$ has already been tested in the beginning, the negative feedback of category $j$ is not repeated by re-allotting it. In lieu of that the policy moves on and a new feedback is obtained at the end which lay be 1 or 0. Thus $\psi$ generates at least as many relevant recommendations as $\psi'$.

Thus the number of relevant products shown under policy $\psi$ is at least as high as that under policy $\psi'$ for every such event $W$ from a set of disjoint events whose union is the entire probability space. Thus the expected number of relevant products shown is also at least as high.

\end{proof}
\subsection{Proof of Theorem~\ref{algo1}}
\begin{proof}
First note that if the relevance matrix $Q$ is such that all the categories form a single non-dominated equivalence class, then the farsighted greedy policy is the same as the optimal policy and so $\overline{W}(Q, p, \beta)=\overline{V}(Q, p, \beta)$. Now consider an experimentation opportunity with an associated relevance matrix $Q$ that has $K$ non-dominated equivalence classes $(U_1,\cdots, U_K)$. Further assume that there is some factor $0<\gamma<1$ such that 
$$\frac{\overline{W}(Q_k^{\pi},p^{\pi}_k,\beta)}{\overline{V}(Q_k^{\pi},p^{\pi}_k,\beta)}\geq \gamma$$
for each $k=1,\cdots, K$, and each $\pi \subset \{1,\cdots,K\}$ such that $k\notin \pi$. Now we have
$$V_k^{\pi}=P(\omega(\pi,k))(\frac{1-\beta^{L^k}}{1-\beta} + \beta^{L^k} \overline{V}(Q_k^{\pi},p^{\pi}_k,\beta))$$ and
\begin{eqnarray}
W^{\pi}_k&=&P(\omega(\pi,k))(\frac{1-\beta^{L^k}}{1-\beta} + \beta^{L^k} \overline{W}(Q_k^{\pi},p^{\pi}_k,\beta))\nonumber\\
&\geq &P(\omega(\pi,k))(\frac{1-\beta^{L^k}}{1-\beta} +\gamma\beta^{L^k} \overline{V}(Q_k^{\pi},p^{\pi}_k,\beta)).
\end{eqnarray}
We thus have
\begin{equation}
\frac{W^{\pi}_k}{V_k^{\pi}}\geq \frac{\frac{1-\beta^{L^k}}{1-\beta} +\gamma \beta^{L^k} \overline{V}(Q_k^{\pi},p^{\pi}_k,\beta)}{\frac{1-\beta^{L^k}}{1-\beta} + \beta^{L^k} \overline{V}(Q_k^{\pi},p^{\pi}_k,\beta)}
\end{equation}
Now it can be easily verified that for a positive constant $c$, the function $f(u)=\frac{c+\gamma u}{c+u}$ is strictly decreasing in $u$. Thus, since $\overline{V}(Q_k^{\pi},p^{\pi}_k,\beta)\leq \frac{1}{1-\beta}$, we have that 
\begin{equation}\label{bound1}
\frac{W^{\pi}_k}{V_k^{\pi}}\geq \frac{\frac{1-\beta^{L^k}}{1-\beta} +\frac{\gamma \beta^{L^k}}{1-\beta}}{\frac{1-\beta^{L^k}}{1-\beta} + \frac{\beta^{L^k}}{1-\beta}}=1-(1-\gamma)\beta^{L^k}\geq 1-(1-\gamma)\beta^{L^{min}}
\end{equation}
where $L^{min}=\min_{k}L^k$. This bound holds uniformly for each $k=1,\cdots, K$, and each $\pi \subset \{1,\cdots,K\}$ such that $k\notin \pi$. Now if we define 
$$OPT'\triangleq$$
\begin{equation}\label{dum}
 \max_{k_1,\cdots,k_{K}\in \sigma(1,\cdots, K)} W_{k_1}+\beta W^{k_1}_{k_2}+\cdots + \beta^{K-1} W^{k_1,\cdots,k_{K-1}}_{k_{K}},
\end{equation}
then from $(\ref{bound1})$, one can easily show that $\frac{OPT'}{V(Q,p,\beta)}\geq  1-(1-\gamma)\beta^{L^{min}}$. Now we will show that the farsighted greedy algorithm attains $\frac{1}{1+\beta-\beta^K}$ factor of $OPT'$. To show this we will use induction in the dynamic programming problem that solves $(\ref{dum})$. Let $\alpha_i$ be the lower bound on the ratio of the payoff to go under the greedy policy and that under the optimal policy when the number of classes left is $i$ where $i$ varies from $1$ to $K$ in the problem $(\ref{dum})$. We are interested in proving that $\alpha_{K}\geq \frac{1}{1+\beta-\beta^K}$. Now if $k_1,\cdots, k_{K-1}$ is decided then there is only one option left for $k_{K}$ and hence the greedy policy gives the same payoff as the optimal payoff to go. Thus $\alpha_1=1$.
Now fix an $i\geq 2$ and consider the payoff to go under the optimal policy when $K-i$ classes in the order have been selected. Let the set of these classes already selected be denoted by labels in $\pi$ and denote this optimal payoff to go by $G^{\pi}_{OPT'}$. Denote the payoff to go under the greedy policy by $G^{\pi}_{g}$. Let the class selected by the greedy policy next be $U_k$ for $k \in \{1,\cdots,K\}\setminus\pi$. Then we have by the definition of $\alpha_{i-1}$:
\begin{equation}\label{main}
G^{\pi}_{g}=W_k^{\pi} +\beta G^{\pi\cup k}_{g} \geq W_k^{\pi} +\beta\alpha_{i-1}G^{\pi\cup k}_{OPT'}
\end{equation}
Now first we have
\begin{equation}
G^{\pi}_{OPT'}=\max_{j\in {1,\cdots,K}\setminus\pi} W_j^{\pi}+\beta G^{\pi\cup j}_{OPT'}.
\end{equation}
Suppose now that a genie reveals the feedback for a class $U_k$ for free at this point. Then the optimal payoff under this new information is higher than the optimal payoff if this information is not available, i.e. $G^{\pi}_{OPT'}$ (because one can always choose to ignore the genie). Denote this optimal payoff under the new information structure as $\bar{G}^{\pi}_{OPT'}$. Then under this new information, clearly if it is revealed that the feedback for $U_k$ is positive then one exhausts all the advertisers in $U_k$, where as if the feedback is negative then one removes $U_k$ from the set of classes and moves on without wasting any opportunity on testing $U_k$. Thus
\begin{equation}
\bar{G}^{\pi}_{OPT'}=W_k^{\pi}+G^{\pi\cup k}_{OPT'} \geq G^{\pi}_{OPT'}.
\end{equation}
And we thus have
\begin{equation}\label{yes}
G^{\pi\cup k}_{OPT'}\geq G^{\pi}_{OPT'}-W_k^{\pi}.
\end{equation}
Substituting $(\ref{yes})$ in $(\ref{main})$ we have
\begin{eqnarray}\label{main2}
G^{\pi}_{g}&\geq& W_k^{\pi} +\beta\alpha_{i-1}(G^{\pi}_{OPT'}-W_k^{\pi})\nonumber\\
&=&W_k^{\pi}(1-\beta\alpha_{i-1}) +\beta\alpha_{i-1} G^{\pi}_{OPT'}
\end{eqnarray}
Further observe that since the greedy policy chooses $k$, we have $G^{\pi}_{OPT'}\leq W_k^{\pi}\frac{1-\beta^{i}}{1-\beta}$ or $W_k^{\pi}\geq G^{\pi}_{OPT'}\frac{1-\beta}{1-\beta^{i}}$ and thus we have
\begin{eqnarray*}
\alpha_i=\frac{G^{\pi}_{g}}{G^{\pi}_{OPT'}}&\geq& \frac{(1-\beta)(1-\beta\alpha_{i-1})}{1-\beta^i}+\beta\alpha_{i-1}\\
&\geq& \frac{(1-\beta)(1-\beta\alpha_{i-1})}{1-\beta^{K}}+\beta\alpha_{i-1}.
\end{eqnarray*}
Here the second inequality follows since $i\leq K$. Now consider the recurrence equation
\begin{equation}
\alpha_i= \frac{(1-\beta)(1-\beta\alpha_{i-1})}{1-\beta^{K}}+\beta\alpha_{i-1}.
\end{equation}
We have that $\alpha_i\leq \alpha_{i-1}$ for $\alpha_{i-1}\geq \frac{1}{1+\beta-\beta^K}$ and hence the sequence $\{\alpha_i\}$ generated by the recurrence relation, with $\alpha_1=1$ is decreasing as long as $\alpha_{i}\geq \frac{1}{1+\beta-\beta^K}$. Further, we can verify that for $\alpha_{i-1}\geq \frac{1}{1+\beta-\beta^K}$, we have
\begin{eqnarray*}
\alpha_{i-1}-\alpha_{i}&=&\alpha_{i-1}(1-\beta)-\frac{(1-\beta)(1-\beta\alpha_{i-1})}{1-\beta^{K}}\\
&\leq& \alpha_{i-1}-\frac{1}{1+\beta-\beta^K}.
\end{eqnarray*}
Thus we can conclude that the sequence $\{\alpha_i\}$ is uniformly bounded below by $\alpha^*=\frac{1}{1+\beta-\beta^K}$ which is the fixed point of the recurrence equation.
Thus we have that $\alpha_{K}\geq \frac{1}{1+\beta-\beta^K}$ which is what we desired to prove.

Thus after combining the bounds, we have that 
\begin{equation}
\frac{\overline{W}(Q,p,\beta)}{\overline{V}(Q,p,\beta)}\geq \frac{1-(1-\gamma)\beta^{L^{min}}}{1+\beta-\beta^K}.
\end{equation}
Let $L_{min}$ be the minimum number of products in any category in $L$, i.e. $L_{min}=\min_{j=1,\cdots,H}L_j$. Now since $L_{min}\leq L^{min}$ and $K\leq H$, which is the total number of categories, we have the following bound that holds irrespective of the relevance matrix $Q$ at any given level:
\begin{equation}
\frac{\overline{W}(Q,p,\beta)}{\overline{V}(Q,p,\beta)}\geq \frac{1-(1-\gamma)\beta^{L_{min}}}{1+\beta-\beta^H}.
\end{equation}
Now let $\gamma_1=1$ and for $i\geq 2$ consider the recurrence equation
\begin{equation}
\gamma_i=\frac{1-(1-\gamma_{i-1})\beta^{L_{min}}}{1+\beta-\beta^H}.
\end{equation}
Now $\gamma_i\leq \gamma_{i-1}$ as long as $\gamma_{i-1}\geq \frac{1-\beta^{L_{min}}}{1+\beta-\beta^H-\beta^{L_{min}}}$. Further we can verify that for $\gamma_{i-1}\geq \frac{1-\beta^{L_{min}}}{1+\beta-\beta^H-\beta^{L_{min}}}$,
\begin{eqnarray*}
\gamma_{i-1}-\gamma_{i}&=&\gamma_{i-1}-\frac{1-(1-\gamma_{i-1})\beta^{L_{min}}}{1+\beta-\beta^H}\\
&\leq& \gamma_{i-1}-\frac{1-\beta^{L_{min}}}{1+\beta-\beta^H-\beta^{L_{min}}}
\end{eqnarray*}
We can thus conclude that the sequence $\{\gamma_i\}$ is uniformly bounded below by $\gamma^*=\frac{1-\beta^{L_{min}}}{1+\beta-\beta^H-\beta^{L_{min}}}$ which is the fixed point of the recurrence relation. Thus for any $(Q,p,\beta)$, 
\begin{equation}
\frac{\overline{W}(Q,p,\beta)}{\overline{V}(Q,p,\beta)}\geq \frac{1-\beta^{L_{min}}}{1+\beta-\beta^H-\beta^{L_{min}}}.
\end{equation}
\qed
\end{proof}

\subsection{Proof of Theorem~\ref{algo2}}
\begin{proof}
Consider the set $(U_1,\cdots, U_K)$ of the non-dominated classes of ad categories at the first experimentation opportunity. From the dynamic programming equation (\ref{dyneqn}) we have
$$V_k^{\pi}=P(\omega(\pi,k))(\frac{1-\beta^{L^k}}{1-\beta} + \beta^{L^k} \overline{V}_k^{\pi}).$$
Here $L^k$ as defined before are the number of ads in class $k$ and $\overline{V}_k^{\pi}$ is the optimal payoff-to-go conditional on the event $E$ given that class $k$ is also used up. We will approximate this payoff by $\mu_k^{\pi}$ defined as
$$\mu_k^{\pi}= P(\omega(\pi,k))(\frac{1-\beta^{L^k}}{1-\beta}).$$
Note that under the greedy policy, $k$ is chosen to maximize $\mu_k^{\pi}$. The ratio of the two quantities is
\begin{equation}
\label{b1}
\frac{\mu_k^{\pi}}{V_k^{\pi}}=\frac{\frac{1-\beta^{L^k}}{1-\beta}}{\frac{1-\beta^{L^k}}{1-\beta} + \beta^{L^k} \overline{V}_k^{\pi}} \geq 1-\beta^{L^k}\geq 1-\beta^{L^{min}}.
\end{equation}
Where the first inequality follows since $\beta^{L^k} \overline{V}_k^{\pi}\leq\frac{\beta^{L^k}}{1-\beta}$ and second follows from the definition of $L^{min}$, since lemma \ref{lem:core} says that the number of ads in a class can only grow. We will later show in an example that for our greedy policy, this bound is tight.
Now the optimal policy finds the best order in which to present the non-dominated equivalence classes which solves the following optimization problem.
\begin{equation}
OPT\triangleq\max_{k_1,\cdots,k_{K}}V_{k_1}+\beta V^{k_1}_{k_2}+\beta^2 V^{k_1,k_2}_{k_3}+\cdots + \beta^{K-1}V^{k_1,\cdots,k_{K-1}}_{k_{K}}.
\end{equation}
Consider instead
\begin{equation}\label{dummy}
OPT'\triangleq \max_{k_1,\cdots,k_{K}}\mu_{k_1}+\beta \mu^{k_1}_{k_2}+\beta^2 \mu^{k_1,k_2}_{k_3}+\cdots + \beta^{K-1}\mu^{k_1,\cdots,k_{K-1}}_{k_{K}}.
\end{equation}
Clearly (\ref{b1}) implies that $\frac{OPT'}{OPT}\geq 1-\beta^{L^{min}}$. Now using the same arguments as that used in the proof of Theorem \ref{algo1} for the optimization problem in definition (\ref{dum}), we can show that the greedy algorithm attains $\frac{OPT'}{1+\beta-\beta^K}$ in $(\ref{dummy})$. Since $K\leq H$, the result follows.

\end{proof}
\subsection{Proof of Theorem~\ref{thm:highbeta}}
\begin{proof}
For a user of type $i$, the total number of products with positive feedback is given by $r_i=\sum_{j=1}^Hq^i_jL_j$.
Thus the expected total number of products with positive feedback is
$$R=\sum_{i=1}^N r_iP_X(i)$$
On the event $W$ that the number of display opportunities is greater than $L$, any policy obtains the full payoff of $R$. Thus its expected payoff is bounded by
$$V_{G}\geq P(W)R=P(C\geq L)R=\beta^{L-1}R.$$
Further the optimal policy cannot attain a payoff greater than $R$. Thus the ratio of the payoff under the any policy and that under the optimal policy is at least $\beta^{L-1}$. 
\end{proof}

\subsection{Recursive computation of Farsighted Greedy}
\begin{algorithm}
{\bf Algorithm 4 (Farsighted greedy)}: Function $[\overline{W}(Q, p, \beta), \overline{A}(Q,p,\beta)]$ where $Q$ is a relevance matrix and $p$ is a probability distribution over user types.
\begin{itemize}
\item If $Q$ is empty, return $\overline{W}(Q,p,\beta)=0$.
\item If $Q$ is non-empty, let the non-dominated equivalence classes be $(U_1,\cdots,U_K)$ and the number of products in each class be denoted by $L^k=\sum_{j\in U_k}L_j$.  Let $N$ be the number of rows in $Q$ corresponding to the user types.
\item Let the event $\omega(k)$ be the event $\{X\in S(k)\}$ where 
$$S(k)=\{i\in N: q^i_j=1\, \forall\, j\in U_k\}.$$ 
Let $Q^k$ be the matrix obtained after removing all the columns corresponding to categories in $U_k$ and the rows corresponding to all the user types in $S(k)^c$ and let $Q_{res}^{k}$ be the matrix obtained after removing all the columns corresponding to categories in  $U_k$ and the rows corresponding to all the user types in $S(k)$. Finally, let $p^k$ denote the distribution on the user types conditional on the event $\{X\in S(k)\}$ and $p^k_{res}$ be the distribution on the user types conditional on $\{X\in S(k)^c\}$.

\item Then define 
 $$W_k=P(S(k))\bigg(\frac{1-\beta^{L^k}}{1-\beta} + \beta^{L^k} \overline{W}(Q^k,p^k,\beta)\bigg) $$
\item Let $W^*=\max_k W_k$ and let $k^*\in \arg\max_kW_k$

Return
 $$\overline{W}(Q,p,\beta)=W^*+\beta(1-P(S(k^*))\overline{W}(Q_{res}^{k^*},p_{res}^{k^*},\beta).$$
$$\overline{A}(Q,p,\beta)=k^*.$$
\end{itemize}
\end{algorithm}

\end{document}